\pdfoutput=1
\documentclass[11pt]{article}

\usepackage[final]{acl}

\usepackage{times}
\usepackage{latexsym}

\usepackage[T1]{fontenc}
\usepackage[utf8]{inputenc}

\usepackage{microtype}

\usepackage{inconsolata}

\usepackage{graphicx}
\usepackage{amssymb}
\usepackage{amsmath}
\usepackage{breqn}
\usepackage{currency}

\DefineCurrency{EUR}{
    name={euro},
    plural={euros},
    symbol={\euro},
    iso={EUR},
    kind=symbol,
}

\title{\textbf{Rosetta-PL: Propositional Logic as a Benchmark for Large Language Model Reasoning}}

\author{
  \textbf{Shaun Baek\textsuperscript{1}\thanks{These authors contributed equally to this work.}, 
Shaun Esua-Mensah\textsuperscript{2}\footnotemark[1], 
Cyrus Tsui\textsuperscript{2}\footnotemark[1], 
Sejan Vigneswaralingam\textsuperscript{2}\footnotemark[1],}\\[1ex]
\textbf{Abdullah Alali\textsuperscript{2},
Michael Lu\textsuperscript{2}\thanks{Lead Investigator}, 
Vasu Sharma\textsuperscript{2}, 
Sean O'Brien\textsuperscript{2},
Kevin Zhu\textsuperscript{2}}\\[1ex]
\textsuperscript{1}Emory University \quad
\textsuperscript{2}Algoverse AI Research\\[1ex]
\texttt{shaun.baek@emory.edu, kevin@algoverseairesearch.org}
}

\begin{document}
\maketitle
\begin{abstract}
Large Language Models (LLMs) are primarily trained on high-resource natural languages, limiting their effectiveness in low-resource settings and in tasks requiring deep logical reasoning. This research introduces Rosetta-PL, a benchmark designed to evaluate LLMs' logical reasoning and generalization capabilities in a controlled environment. We construct Rosetta-PL by translating a dataset of logical propositions from Lean into a custom logical language, which is then used to fine-tune an LLM (e.g., GPT-4o). Our experiments analyze the impact of the size of the dataset and the translation methodology on the performance of the model. Our results indicate that preserving logical relationships in the translation process significantly boosts precision, with accuracy plateauing beyond roughly 20,000 training samples. These insights provide valuable guidelines for optimizing LLM training in formal reasoning tasks and improving performance in various low-resource language applications.
\end{abstract}

\section{Introduction}

Large Language Models (LLMs), such as OpenAI's GPT models \citep{brown2020languagemodelsfewshotlearners}, Google's Gemini models \citep{geminiteam2024geminifamilyhighlycapable}, and Meta's Llama models \citep{touvron2023llama2openfoundation}, are typically trained on high-resource natural languages (e.g., English, Spanish, and Chinese). This focus on high-resource languages disadvantages speakers of low-resource languages, as training models for these languages are more challenging due to their inherent complexity \citep{nllbteam2022languageleftbehindscaling}. Furthermore, semantic ambiguity, grammatical complexities, and contextual dependencies in natural languages can limit the capabilities of an LLM in precise logical reasoning. Since natural language often relies on implied meaning, subtle cues, and flexible syntax, models trained primarily on data using these principles may struggle to follow strict rules needed for logical reasoning \citep{asher2023limitslearninglanguagemodels}.

To isolate these reasoning abilities from language-specific challenges, we propose the evaluation of LLMs within a controlled setting using formal logical language. Logical languages, characterized by strict syntax and precise semantics, eliminate many of the extraneous factors present in natural languages, allowing us to focus squarely on pattern recognition and problem solving. Although prior benchmarks, such as LOGIGLUE \citep{luo2024logigluebriefsurveybenchmark}, provide structured reasoning tasks, these typically rely on predefined reasoning steps, making it challenging to determine whether an LLM can autonomously identify and apply logical rules. In contrast, our benchmark, Rosetta-PL, evaluates whether LLMs can discover logical patterns within a propositional language, thereby measuring reasoning ability without relying on predefined inference steps or extraneous linguistic factors. Research on applying LLMs to logic-based problem solving is relatively scarce, and while chain-of-thought (CoT) prompting has gained popularity in natural language tasks \citep{wei2023chainofthoughtpromptingelicitsreasoning}, its effectiveness in logical or symbolic contexts remains largely unexplored \citep{creswell2022selectioninferenceexploitinglargelanguage}. 

We address this gap by constructing Rosetta-PL by translating the Lean Workbook dataset \citep{ying2024leanworkbooklargescalelean} into our own propositional language and fine-tuning ChatGPT \citep{brown2020languagemodelsfewshotlearners} using the translated dataset. We evaluate logical accuracy in our custom language while varying training data parameters such as training set size and the method of translation. Our experiments point towards potentially effective training strategies and provide preliminary estimates on the dataset size needed to approach benchmark-level logical understanding. By setting aside language-specific factors, we focus on the relationship between pattern recognition and data requirements, offering insights that impact language training in both high- and low-resource settings.

\section{Background}
Large Language Models (LLMs) have excelled at tasks involving unstructured natural language, yet their capacity for structured logical reasoning remains underexplored \citep{creswell2022selectioninferenceexploitinglargelanguage}. The inherent ambiguities of natural language, such as polysemy and idiomatic expressions, can obscure true reasoning capabilities. In contrast, formal logical languages, defined by strict syntax and unambiguous semantics, offer a controlled testbed for evaluating pattern recognition and rule-based inference \citep{barcelo2023logicallanguagesacceptedtransformer}.

Propositional logic, a fundamental component of formal logic, employs connectives (e.g., $\land$, $\lor$, $\neg$) to combine atomic propositions into complex expressions whose truth values are fully determined by their parts \citep{niu2024grobnershirshovbasesmarkovsemirings}. This clarity makes it an ideal framework for assessing whether LLMs can autonomously learn and generalize logical rules—a skill central to disciplines like mathematics and programming \citep{nye2021workscratchpadsintermediatecomputation, polu2020generativelanguagemodelingautomated}.

Recent benchmarks have begun to probe the symbolic reasoning of LLMs. For example, LOGIC-LM demonstrates that LLMs can solve logic puzzles when aided by external symbolic solvers \citep{pan2023logiclmempoweringlargelanguage}. Meanwhile, LOGIGLUE \citep{luo2024logigluebriefsurveybenchmark} and Logic Bench \citep{parmar2024logicbenchsystematicevaluationlogical} evaluate multi-step reasoning based on predefined inference templates, and chain-of-thought prompting has been shown to improve arithmetic performance \citep{wei2023chainofthoughtpromptingelicitsreasoning}. Other studies have further enriched this landscape: for example, the SymbCoT framework integrates symbolic expressions and logic rules directly into chain-of-thought (CoT) thereby boosting reasoning fidelity \citep{xu2024faithfullogicalreasoningsymbolic}, while research examining the impact of symbolic solver choices has revealed that tool selection (e.g., Z3, Prover9, or Pyke) can cause performance variations of up to 50\% \citep{lam2024closerlooklogicalreasoning}. Furthermore, work on step-by-step symbolic verification has demonstrated that automated checks of intermediate reasoning steps can substantially enhance overall accuracy \citep{zhang2024evaluatingstepbystepreasoningsymbolic}. However, these approaches tend to rely on surface-level statistical correlations rather than genuine discovery of novel logical patterns \citep{creswell2022selectioninferenceexploitinglargelanguage}.

To bridge this gap, our work translates natural language logic problems into a propositional language, thereby eliminating linguistic complexities and focusing solely on intrinsic pattern recognition. Building on formal frameworks such as Lean4 \citep{ying2024leanworkbooklargescalelean}, we investigate how well LLMs can learn and generalize new logical structures—a capability that also carries implications for improving training strategies in low-resource language settings \citep{nllbteam2022languageleftbehindscaling}.

\section{Method}
\subsection{Objective}
The primary objective of this experiment is to evaluate the logical accuracy and pattern recognition capabilities of LLMs in a newly created propositional language. By removing linguistic complexities to focus solely on logical problem-solving, we aim to determine how well these models generalize and adapt in a structured, logic-based environment under varying dataset sizes, and whether this process reveals or rectifies discrepancies in their understanding of formal languages.

\subsection{Dataset}

We derived Rosetta-PL from the Lean Workbook \citep{ying2024leanworkbooklargescalelean}, which is a dataset of logical problems translated into the formal language of Lean. Each problem was translated into our custom propositional language using a predefined translation key, resulting in a training dataset of 25,214 problems. Each dataset entry was written in a conversation-like structure with system, user, assistant, function, and message content, containing a logical problem (a statement) in our custom language and its corresponding truth value, indicating whether the statement is true or not. In contrast to benchmarks such as LOGIGLUE \citep{luo2024logigluebriefsurveybenchmark} and LOGIC-LM \citep{pan2023logiclmempoweringlargelanguage}, which focus on logical problems with predefined inference steps, Rosetta-PL is designed to test an LLM's ability to discover new patterns. Unlike Logic Bench \citep{parmar2024logicbenchsystematicevaluationlogical}, which evaluates performance on known logical patterns, our dataset requires the model to infer novel patterns.

\subsection{Experimental Methodology}

 Our experimental setup involved building a data pipeline for fine-tuning GPT-4o on formal logical tasks.  We opted to use GPT-4o primarily due to its performance on a range of reasoning benchmarks such as MMLU (Massive Multitask Language Understanding), GSM8K, and Big Bench Hard, allowing us to compare with one of the highest performers for LLMs in formal logic tasks. Because GPT-4o is closed-source, there is an inherent risk of leakage challenges. However, by translating the Lean Workbook into our own custom propositional language, we altered the original problems in an unorthodox way that makes direct overlap in GPT-4o’s training far less likely.

Each entry in our training dataset was verified to conform to the required format—ensuring valid roles such as system, user, and assistant, and is passed on to GPT-4o for fine-tuning. From this same dataset, we also extracted "seen" testing subsets by randomly selecting 500 entries. We also extracted "unseen" testing subsets by randomly selecting 200 problems from an entirely different source: the Minif2f-lean4 dataset \citep{zheng2022minif2fcrosssystembenchmarkformal}, which does not overlap with the training dataset. We aim to measure the model's ability to both retain learned information and generalize its logical understanding to novel patterns through the "seen" and "unseen" datasets respectively.
 
 Throughout these experiments, all fine-tuning and testing were conducted using NVIDIA A100 GPUs. Overall, GPT-4o underwent four separate fine-tuning runs, during which we kept parameter settings constant (e.g., learning rate, number of epochs) while varying the size of the training dataset (25,214, 20,000, and 10,000) and which one out of the two translation keys used. These translation keys altered how the logical problems from the Lean Workbook were mapped into our custom language, effectively creating multiple languages with varying logical structures.

Original Example: 
\newcommand*{\field}[1]{\mathbb{#1}}%
\begin{dmath}
  x y z : \field{N} \vdash (x ^ 2 + 1) * (y ^ 2 + 1) * (z ^ 2 + 1) = (x + y + z) ^ 2 - 2 * (x * y + y * z + z * x) + (x * y + y * z + z * x) ^ 2 - 2 * x * y * z * (x + y + z) + x ^ 2 * y ^ 2 * z ^ 2 + 1
\end{dmath}

 \paragraph{Translation Strategies:}

To investigate the effect of symbolic representation on logical reasoning, we employ two distinct translation strategies. The first strategy maintains the inherent logical relationships by carefully mapping symbols, while the second intentionally disrupts these patterns through arbitrary transformations. These contrasting approaches allow us to assess how preserving or altering logical structure influences model performance.

\begin{itemize}
    \item Translation Key 1 Strategy (Focused Key): Translation Key 1 replaces Lean symbols with other symbols (see appendix). This method preserves logical relationships by ensuring that related symbols are consistently mapped. For instance, the symbols “>” and “<” are translated into “>>” and “<<”, respectively, preserving their comparative meaning. This is to mimic spoken language, where symbols and phrases are logically related. Additionally, the sentence structure is encrypted using a scrambling function that adds a reversed duplicate of the sentence at the end, with a few additional symbols in between, in order to mimic the variations in sentence structures across different languages. An example of an entry translated with Key 1 is shown below:
\end{itemize}
\begin{dmath}
    x y z \neg \field{N} \#\# |-|-|-x \wedge \allowbreak \wedge 2 \wedge \wedge 1|- $\mbox{\euro}$ |-|-|-y \wedge \wedge 2 \wedge \wedge 1|-\allowbreak  $\mbox{\euro}$ |-|-|-z \wedge \wedge 2 \wedge \wedge 1|- == |-|-|-x \wedge \wedge y \wedge \wedge z| \allowbreak - \wedge \wedge 2 _ 2 $\mbox{\euro}$ |-|-|-x $\mbox{\euro}$  y \wedge \wedge y $\mbox{\euro}$ z \wedge \wedge z € x|- \wedge \wedge |-|-|-x $\mbox{\euro}$ y \allowbreak \wedge \wedge y  $\mbox{\euro}$ z \wedge \wedge z $\mbox{\euro}$ x|- \wedge \wedge 2 _ 2 $\mbox{\euro}$ x $\mbox{\euro}$ y $\mbox{\euro}$ z $\mbox{\euro}$ |-|-|-x \wedge \wedge y \wedge \wedge z| \allowbreak - \wedge \wedge x \wedge \wedge 2  $\mbox{\euro}$ y \wedge \wedge 2 $\mbox{\euro}$ z \wedge \wedge 2 \wedge \wedge 1
\end{dmath}

\begin{itemize}
    \item Translation Key 2 Strategy (Random Key): In contrast, this method removes logical structure by shifting the ASCII values of each character by 10, resulting in an entirely arbitrary transformation. As a result, the translated expression loses any recognizable logical patterns. Additionally, statements are inverted around logical operators such as ->, >, <, >=, and <=. For example, an expression of the form “A > B > C” would be translated into “C T(>) B T(>) A”, where T(>) represents the transformed version of the “>” symbol. An example of an entry translated with Key 2 is provided below:
\end{itemize}
\begin{dmath}
\verb|"y!z!{!;!\u2125\u000b\u22a3!)y!_!3!,| \allowbreak
\verb|!2*!+!)z!_!3!,!2*!+!){!_!3!,!2*!>\u000b|\allowbreak
\verb|!!!!)y!,!z!,!{*!_!3!.!3!+!)y!+!z!,!z!+!|\allowbreak
\verb|{!,!{!+!y*!,!)y!+!z!,!z!+!{!,!{!+!y*!_|\allowbreak
\verb|!3!.!3!+!y!+!z!+!{!+!)y!,!z!,!{*!,|\allowbreak
\verb|\u000b!!!!!!!!y!_!3!+!z!_!3!+!|\allowbreak
\verb|!,\u000b!!!!!!2"||\allowbreak
\end{dmath}

 \paragraph{Evaluation Procedure:}

We conducted four fine-tuning runs on GPT-4o, keeping all hyperparameters constant, and evaluated five models (four fine-tuned and one base model with no fine-tuning) using 12 distinct datasets. These datasets are organized into two main categories:

\begin{itemize}
    \item Seen Data: Six datasets were created by randomly selecting problems from the training set—three datasets containing 500 problems each in the original Lean format and three datasets with 500 problems each using the same translation key employed during fine-tuning.
    \item Unseen Data: To assess generalization, six additional datasets were formed by randomly selecting 200 problems each from the independent Mini-f2f dataset (Zheng et al., 2022). Like the seen data, these were split into two groups of three datasets: one in Lean and the other using the corresponding translated format.
\end{itemize}

Overall accuracy was computed by averaging the results across all testing sets, with accuracy defined as the number of correctly answered queries divided by the total number of queries in each set.

\section{Results}
Figure 1 displays the comparative performance of four fine-tuned GPT-4o models evaluated on both “seen” and “unseen” datasets. Specifically, models were fine-tuned with 25,214, 20,000, and 10,000 distinct queries using Translation Key 1, and with 25,214 queries using Translation Key 2. Additionally, Lean (untranslated) versions of both testing sets serve as benchmarks.

\begin{figure*}
    \centering
    \includegraphics[width=\textwidth]{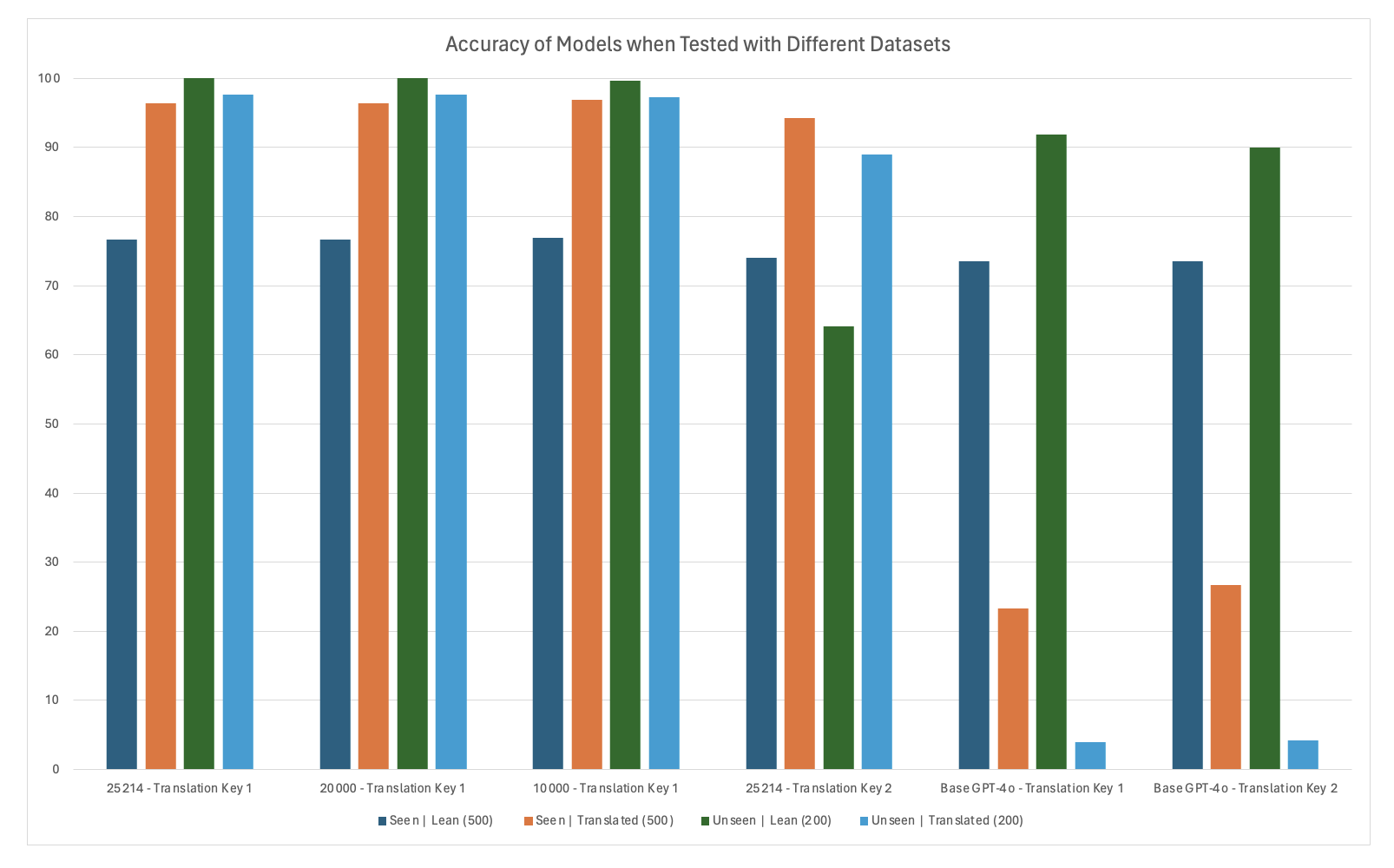}
    \caption{Comparison of GPT-4o accuracy across datasets ("Seen" and "Unseen") using different translation keys and varying dataset sizes.}
    
\end{figure*}

Our experiments demonstrate that GPT-4o exhibits superior problem-solving performance in our custom propositional language compared to Lean on average. On the “seen” dataset, GPT-4o achieved an average accuracy over all tests in of 95.97\% in our propositional language versus 76.08\% in Lean, with a small uncertainty of $\pm$ 0.33\% and $\pm$ 0.36\% respectively.

In contrast, on the “unseen” dataset, GPT-4o performed better when tested in Lean than in our custom language—attaining 99.89\% accuracy with Lean compared to 97.56\% with Translation Key 1 ($\pm$ 0.06\% and $\pm$ 0.44\% respectively). As expected, Translation Key 2 yielded a substantially lower accuracy of 64.1\% ($\pm$ 0.75\%) due to its arbitrary mapping. The model was fine-tuned solely on translated data, so it specializes in those patterns, resulting in high performance on seen translated examples but poor performance on seen Lean examples. For unseen data, it falls back on its broader pre-training, which helps it perform better on unseen Lean problems.

Additionally, our experiments indicate that GPT-4o solves problems more accurately with Translation Key 1 than with Translation Key 2, with average accuracies of 92.68\% compared to 80.36\% respectively—highlighting the importance of preserving logical relationships in the translation process. Table \ref{tab:performance-translationkey1} provides a detailed summary of results from testing with Translation Key 1, and Table \ref{tab:performance-translationkey2} provides a detailed summary of results from testing with Translation Key 2.

Furthermore, training set size influenced performance. Increasing the training set from 10,000 to 20,000 samples improved accuracy by 2.7\% on the “seen” dataset and by 0.3\% on the “unseen” dataset, while further increases up to 25,214 samples did not yield additional gains. This suggests that the training set size threshold for stable performance lies below 20,000 samples.

For seen data in the custom translated format, the fine-tuned GPT-4o consistently achieves higher accuracy by specializing in the patterns and syntax introduced during fine-tuning, outperforming the base model. In contrast, on seen Lean data, the base GPT-4o retains its general Lean knowledge from pre-training and achieves similar results to the fine-tuned model.

When it comes to unseen data, the fine-tuned GPT-4o expectedly outperforms the base model on unseen translated examples. Table \ref{tab:best-gpt4o} provides a detailed summary of the results from testing using the base GPT-4o model. However, for unseen Lean data, the GPT-4o fine-tuned using Translation Key 2 performed significantly worse than its Translation Key 1 counterparts and also the base models. Focusing on Lean data (untranslated), all 4 fine-tuned models outperform the base models in both the unseen and seen data, except for the model fine-tuned in Translation Key 2 which showed worse comparative performance in the unseen lean data.

Tables \ref{tab:performance-translationkey1}, \ref{tab:performance-translationkey2}, and \ref{tab:best-gpt4o} provides a detailed summary of all dataset permutations and average performance metrics, shedding light on any potential anomalies.

\section{Discussion}
Our findings align with previous studies \citep{kojima2023largelanguagemodelszeroshot, wei2023chainofthoughtpromptingelicitsreasoning}, demonstrating that the accuracy of logical reasoning depends significantly on prompt formulation and task representation. The use of translation keys in our experiments illustrates that preserving inherent logical relationships—as in Translation Key 1—yields better performance than employing arbitrary mappings. This is analogous to natural language, where inverse or comparable relationships between symbols facilitate comprehension.

Our results also reveal a general trend where accuracy increases with training set size, echoing prior research that shows LLMs can perform well even with limited data \citep{brown2020languagemodelsfewshotlearners}. However, as shown in Figure 1, this trend is not strictly linear. There are occasions where smaller datasets outperformed larger datasets, such as the "seen" dataset in our propositional language having a 0.467\% greater accuracy with 10000 samples compared to 20,000 samples. We attribute these fluctuations to certain factors, such as overfitting in larger training sets. Unlike earlier studies that evaluated existing models \citep{liu2023gloreevaluatinglogicalreasoning}, our approach using a custom propositional language uncovers unique aspects of pattern recognition in LLMs.

Notably, our analysis revealed that GPT-4o's performance on unseen data is better in Lean than it is in our custom language. We attribute this to GPT-4o’s prior exposure to Lean-like syntax during pre-training Lean, as a formal proof assistant, shares structural similarities with theorem-proving and programming languages. 
In contrast, the custom language, especially under Translation Key 2, disrupted logical structure, thereby impeding generalization. This suggests that fine-tuning benefits significantly when the training data preserves logical consistency, aligning with the model's pre-training experience. 

This is further reinforced by the observation that models fine-tuned with Translation Key 1 performed better across all testing sets than those fine-tuned with Translation Key 2. Additionally, the fine-tuned models—especially those with Translation Key 1—consistently exhibited superior performance on both seen and unseen data, and this performance improved with larger training set sizes. This demonstrates GPT's ability to generalize logical information. The LLM extracted logical information from our custom language and used it to improve its logical accuracy in Lean. Notably, it performed better with Translation Key 1—which preserves logical relationships—than with Translation Key 2, which disrupts them.

While distinguishing between these effects is challenging, future work could explore fine-tuning an LLM with minimal exposure to Lean syntax to better understand the impact of pre-training familiarity compared to logical structure preservation. Comparing performance across runs provided insights into whether GPT-4o could robustly handle shifts in symbolic representation and how sensitive its performance is to different training configurations.

Our experiments indicate that GPT-4o’s performance plateaus at around 20,000 training examples. This plateau may result from dataset redundancy, model capacity limitations, or the relative simplicity of the tasks. When the dataset contains many similar patterns, the model's exposure to novel challenges is limited, and once key patterns are internalized, additional training yields diminishing returns.

In summary, our findings suggest that GPT-4o can achieve high problem-solving accuracy in a propositional language when fine-tuned appropriately. The choice of translation key, dataset characteristics, and training set size must be managed carefully to mitigate overfitting and ensure robust generalization beyond seen patterns. 

\section{Conclusion}
Our investigation confirms that fine-tuning GPT-4o on a custom propositional language not only facilitates high-level logical reasoning but also underscores the critical role of maintaining relational integrity within training data. Specifically, our work shows that using structured translation strategies significantly enhances model performance. This improvement is achieved by aligning the training data with the inherent logical patterns familiar from the model’s pre-training, allowing GPT-4o to generalize more effectively, particularly when transitioning from seen to unseen examples.

Furthermore, our analysis highlights that an optimally balanced training set is essential: while increased dataset size improves performance up to a threshold (around 20,000 examples), additional data yields diminishing returns, suggesting the need for more efficient data utilization methods. These findings not only validate the importance of structured prompts and contextual cues but also offer practical guidelines for optimizing LLM training in both high- and low-resource language scenarios. 

Collectively, our results contribute to a deeper understanding of how targeted data curation and translation methodologies can bolster logical reasoning in large language models.

\section{Future Research}
Future work should investigate dataset design principles. The high accuracy observed on our unseen dataset may reflect biases, such as overrepresentation of certain problem types or cultural premises, which should be systematically addressed. Synthetically balanced datasets that incorporate tiered complexity levels (e.g., single-step versus multi-step reasoning) could help disentangle superficial pattern recognition from genuine logical understanding. Additionally, although formatting differences (e.g., brackets versus colons) did not hinder performance in our study, systematic evaluations of robustness to syntactic variations are needed to better assess adaptability in low-resource settings.

A potential path to explore would be foregoing fine-tuning GPT-4o on our custom dataset and instead rely on in-context learning. Because GPT-4o may already have some familiarity with Lean from its pre-training, one could design a prompt that includes a few worked examples of Lean problems alongside a call to an external translator function that converts Lean input into the custom propositional language at inference time. Though this may yield lower accuracy than fine-tuning, it avoids the cost of creating and maintaining a large translation corpus. Evaluating GPT-4o in context can reveal how much of its Lean knowledge can be utilized through prompt engineering alone.

Further research should focus on optimizing translation strategies by developing principled approaches, such as semantic alignment of symbols, to enhance learnability. At the same time, exploring data efficiency methods is critical, as our observed performance plateau at approximately 20,000 training examples suggests that smarter data utilization may both reduce data requirements and improve systematicity.

\clearpage

\bibliography{algoverse}

\begin{thebibliography}{21}
\providecommand{\natexlab}[1]{#1}

\bibitem[{Asher et~al.(2023)Asher, Bhar, Chaturvedi, Hunter, and Paul}]{asher2023limitslearninglanguagemodels}
Nicholas Asher, Swarnadeep Bhar, Akshay Chaturvedi, Julie Hunter, and Soumya Paul. 2023.
\newblock \href {https://arxiv.org/abs/2306.12213} {Limits for learning with language models}.
\newblock \emph{Preprint}, arXiv:2306.12213.

\bibitem[{Barcelo et~al.(2023)Barcelo, Kozachinskiy, Lin, and Podolskii}]{barcelo2023logicallanguagesacceptedtransformer}
Pablo Barcelo, Alexander Kozachinskiy, Anthony~Widjaja Lin, and Vladimir Podolskii. 2023.
\newblock \href {https://arxiv.org/abs/2310.03817} {Logical languages accepted by transformer encoders with hard attention}.
\newblock \emph{Preprint}, arXiv:2310.03817.

\bibitem[{Brown et~al.(2020)Brown, Mann, Ryder, Subbiah, Kaplan, Dhariwal, Neelakantan, Shyam, Sastry, Askell, Agarwal, Herbert-Voss, Krueger, Henighan, Child, Ramesh, Ziegler, Wu, Winter, Hesse, Chen, Sigler, Litwin, Gray, Chess, Clark, Berner, McCandlish, Radford, Sutskever, and Amodei}]{brown2020languagemodelsfewshotlearners}
Tom~B. Brown, Benjamin Mann, Nick Ryder, Melanie Subbiah, Jared Kaplan, Prafulla Dhariwal, Arvind Neelakantan, Pranav Shyam, Girish Sastry, Amanda Askell, Sandhini Agarwal, Ariel Herbert-Voss, Gretchen Krueger, Tom Henighan, Rewon Child, Aditya Ramesh, Daniel~M. Ziegler, Jeffrey Wu, Clemens Winter, and 12 others. 2020.
\newblock \href {https://arxiv.org/abs/2005.14165} {Language models are few-shot learners}.
\newblock \emph{Preprint}, arXiv:2005.14165.

\bibitem[{Creswell et~al.(2022)Creswell, Shanahan, and Higgins}]{creswell2022selectioninferenceexploitinglargelanguage}
Antonia Creswell, Murray Shanahan, and Irina Higgins. 2022.
\newblock \href {https://arxiv.org/abs/2205.09712} {Selection-inference: Exploiting large language models for interpretable logical reasoning}.
\newblock \emph{Preprint}, arXiv:2205.09712.

\bibitem[{Kojima et~al.(2023)Kojima, Gu, Reid, Matsuo, and Iwasawa}]{kojima2023largelanguagemodelszeroshot}
Takeshi Kojima, Shixiang~Shane Gu, Machel Reid, Yutaka Matsuo, and Yusuke Iwasawa. 2023.
\newblock \href {https://arxiv.org/abs/2205.11916} {Large language models are zero-shot reasoners}.
\newblock \emph{Preprint}, arXiv:2205.11916.

\bibitem[{Lam et~al.(2024)Lam, Thatikonda, and Shareghi}]{lam2024closerlooklogicalreasoning}
Long Hei~Matthew Lam, Ramya~Keerthy Thatikonda, and Ehsan Shareghi. 2024.
\newblock \href {https://arxiv.org/abs/2406.00284} {A closer look at logical reasoning with llms: The choice of tool matters}.
\newblock \emph{Preprint}, arXiv:2406.00284.

\bibitem[{liu et~al.(2023)liu, Teng, Ning, Liu, Zhou, and Zhang}]{liu2023gloreevaluatinglogicalreasoning}
Hanmeng liu, Zhiyang Teng, Ruoxi Ning, Jian Liu, Qiji Zhou, and Yue Zhang. 2023.
\newblock \href {https://arxiv.org/abs/2310.09107} {Glore: Evaluating logical reasoning of large language models}.
\newblock \emph{Preprint}, arXiv:2310.09107.

\bibitem[{Luo et~al.(2024)Luo, Kumbhar, shen, Parmar, Varshney, Banerjee, Aditya, and Baral}]{luo2024logigluebriefsurveybenchmark}
Man Luo, Shrinidhi Kumbhar, Ming shen, Mihir Parmar, Neeraj Varshney, Pratyay Banerjee, Somak Aditya, and Chitta Baral. 2024.
\newblock \href {https://arxiv.org/abs/2310.00836} {Towards logiglue: A brief survey and a benchmark for analyzing logical reasoning capabilities of language models}.
\newblock \emph{Preprint}, arXiv:2310.00836.

\bibitem[{Niu et~al.(2024)Niu, Li, and Wang}]{niu2024grobnershirshovbasesmarkovsemirings}
Xiaohui Niu, Wenxi Li, and Zhongzhi Wang. 2024.
\newblock \href {https://arxiv.org/abs/2401.05731} {On grobner-shirshov bases for markov semirings}.
\newblock \emph{Preprint}, arXiv:2401.05731.

\bibitem[{Nye et~al.(2021)Nye, Andreassen, Gur-Ari, Michalewski, Austin, Bieber, Dohan, Lewkowycz, Bosma, Luan, Sutton, and Odena}]{nye2021workscratchpadsintermediatecomputation}
Maxwell Nye, Anders~Johan Andreassen, Guy Gur-Ari, Henryk Michalewski, Jacob Austin, David Bieber, David Dohan, Aitor Lewkowycz, Maarten Bosma, David Luan, Charles Sutton, and Augustus Odena. 2021.
\newblock \href {https://arxiv.org/abs/2112.00114} {Show your work: Scratchpads for intermediate computation with language models}.
\newblock \emph{Preprint}, arXiv:2112.00114.

\bibitem[{Pan et~al.(2023)Pan, Albalak, Wang, and Wang}]{pan2023logiclmempoweringlargelanguage}
Liangming Pan, Alon Albalak, Xinyi Wang, and William~Yang Wang. 2023.
\newblock \href {https://arxiv.org/abs/2305.12295} {Logic-lm: Empowering large language models with symbolic solvers for faithful logical reasoning}.
\newblock \emph{Preprint}, arXiv:2305.12295.

\bibitem[{Parmar et~al.(2024)Parmar, Patel, Varshney, Nakamura, Luo, Mashetty, Mitra, and Baral}]{parmar2024logicbenchsystematicevaluationlogical}
Mihir Parmar, Nisarg Patel, Neeraj Varshney, Mutsumi Nakamura, Man Luo, Santosh Mashetty, Arindam Mitra, and Chitta Baral. 2024.
\newblock \href {https://arxiv.org/abs/2404.15522} {Logicbench: Towards systematic evaluation of logical reasoning ability of large language models}.
\newblock \emph{Preprint}, arXiv:2404.15522.

\bibitem[{Polu and Sutskever(2020)}]{polu2020generativelanguagemodelingautomated}
Stanislas Polu and Ilya Sutskever. 2020.
\newblock \href {https://arxiv.org/abs/2009.03393} {Generative language modeling for automated theorem proving}.
\newblock \emph{Preprint}, arXiv:2009.03393.

\bibitem[{Team et~al.(2024)Team, Anil, Borgeaud, Alayrac, Yu, Soricut, Schalkwyk, Dai, Hauth, Millican, Silver, Johnson, Antonoglou, Schrittwieser, Glaese, Chen, Pitler, Lillicrap, Lazaridou, Firat, Molloy, Isard, Barham, Hennigan, Lee, Viola, Reynolds, Xu, Doherty, Collins, Meyer, Rutherford, Moreira, Ayoub, Goel, Krawczyk, Du, Chi, Cheng, Ni, Shah, Kane, Chan, Faruqui, Severyn, Lin, Li, Cheng, Ittycheriah, Mahdieh, Chen, Sun, Tran, Bagri, Lakshminarayanan, Liu, Orban, Güra, Zhou, Song, Boffy, Ganapathy, Zheng, Choe, Ágoston Weisz, Zhu, Lu, Gopal, Kahn, Kula, Pitman, Shah, Taropa, Merey, Baeuml, Chen, Shafey, Zhang, Sercinoglu, Tucker, Piqueras, Krikun, Barr, Savinov, Danihelka, Roelofs, White, Andreassen, von Glehn, Yagati, Kazemi, Gonzalez, Khalman, Sygnowski, Frechette, Smith, Culp, Proleev, Luan, Chen, Lottes, Schucher, Lebron, Rrustemi, Clay, Crone, Kocisky, Zhao, Perz, Yu, Howard, Bloniarz, Rae, Lu, Sifre, Maggioni, Alcober, Garrette, Barnes, Thakoor, Austin, Barth-Maron, Wong, Joshi, Chaabouni,
  Fatiha, Ahuja, Tomar, Senter, Chadwick, Kornakov, Attaluri, Iturrate, Liu, Li, Cogan, Chen, Jia, Gu, Zhang, Grimstad, Hartman, Garcia, Pillai, Devlin, Laskin, de~Las~Casas, Valter, Tao, Blanco, Badia, Reitter, Chen, Brennan, Rivera, Brin, Iqbal, Surita, Labanowski, Rao, Winkler, Parisotto, Gu, Olszewska, Addanki, Miech, Louis, Teplyashin, Brown, Catt, Balaguer, Xiang, Wang, Ashwood, Briukhov, Webson, Ganapathy, Sanghavi, Kannan, Chang, Stjerngren, Djolonga, Sun, Bapna, Aitchison, Pejman, Michalewski, Yu, Wang, Love, Ahn, Bloxwich, Han, Humphreys, Sellam, Bradbury, Godbole, Samangooei, Damoc, Kaskasoli, Arnold, Vasudevan, Agrawal, Riesa, Lepikhin, Tanburn, Srinivasan, Lim, Hodkinson, Shyam, Ferret, Hand, Garg, Paine, Li, Li, Giang, Neitz, Abbas, York, Reid, Cole, Chowdhery, Das, Rogozińska, Nikolaev, Sprechmann, Nado, Zilka, Prost, He, Monteiro, Mishra, Welty, Newlan, Jia, Allamanis, Hu, de~Liedekerke, Gilmer, Saroufim, Rijhwani, Hou, Shrivastava, Baddepudi, Goldin, Ozturel, Cassirer, Xu, Sohn, Sachan,
  Amplayo, Swanson, Petrova, Narayan, Guez, Brahma, Landon, Patel, Zhao, Villela, Wang, Jia, Rahtz, Giménez, Yeung, Keeling, Georgiev, Mincu, Wu, Haykal, Saputro, Vodrahalli, Qin, Cankara, Sharma, Fernando, Hawkins, Neyshabur, Kim, Hutter, Agrawal, Castro-Ros, van~den Driessche, Wang, Yang, yiin Chang, Komarek, McIlroy, Lučić, Zhang, Farhan, Sharman, Natsev, Michel, Bansal, Qiao, Cao, Shakeri, Butterfield, Chung, Rubenstein, Agrawal, Mensch, Soparkar, Lenc, Chung, Pope, Maggiore, Kay, Jhakra, Wang, Maynez, Phuong, Tobin, Tacchetti, Trebacz, Robinson, Katariya, Riedel, Bailey, Xiao, Ghelani, Aroyo, Slone, Houlsby, Xiong, Yang, Gribovskaya, Adler, Wirth, Lee, Li, Kagohara, Pavagadhi, Bridgers, Bortsova, Ghemawat, Ahmed, Liu, Powell, Bolina, Iinuma, Zablotskaia, Besley, Chung, Dozat, Comanescu, Si, Greer, Su, Polacek, Kaufman, Tokumine, Hu, Buchatskaya, Miao, Elhawaty, Siddhant, Tomasev, Xing, Greer, Miller, Ashraf, Roy, Zhang, Ma, Filos, Besta, Blevins, Klimenko, Yeh, Changpinyo, Mu, Chang, Pajarskas, Muir,
  Cohen, Lan, Haridasan, Marathe, Hansen, Douglas, Samuel, Wang, Austin, Lan, Jiang, Chiu, Lorenzo, Sjösund, Cevey, Gleicher, Avrahami, Boral, Srinivasan, Selo, May, Aisopos, Hussenot, Soares, Baumli, Chang, Recasens, Caine, Pritzel, Pavetic, Pardo, Gergely, Frye, Ramasesh, Horgan, Badola, Kassner, Roy, Dyer, Campos, Tomala, Tang, Badawy, White, Mustafa, Lang, Jindal, Vikram, Gong, Caelles, Hemsley, Thornton, Feng, Stokowiec, Zheng, Thacker, Çağlar Ünlü, Zhang, Saleh, Svensson, Bileschi, Patil, Anand, Ring, Tsihlas, Vezer, Selvi, Shevlane, Rodriguez, Kwiatkowski, Daruki, Rong, Dafoe, FitzGerald, Gu-Lemberg, Khan, Hendricks, Pellat, Feinberg, Cobon-Kerr, Sainath, Rauh, Hashemi, Ives, Hasson, Noland, Cao, Byrd, Hou, Wang, Sottiaux, Paganini, Lespiau, Moufarek, Hassan, Shivakumar, van Amersfoort, Mandhane, Joshi, Goyal, Tung, Brock, Sheahan, Misra, Li, Rakićević, Dehghani, Liu, Mittal, Oh, Noury, Sezener, Huot, Lamm, Cao, Chen, Mudgal, Stella, Brooks, Vasudevan, Liu, Chain, Melinkeri, Cohen, Wang,
  Seymore, Zubkov, Goel, Yue, Krishnakumaran, Albert, Hurley, Sano, Mohananey, Joughin, Filonov, Kępa, Eldawy, Lim, Rishi, Badiezadegan, Bos, Chang, Jain, Padmanabhan, Puttagunta, Krishna, Baker, Kalb, Bedapudi, Kurzrok, Lei, Yu, Litvin, Zhou, Wu, Sobell, Siciliano, Papir, Neale, Bragagnolo, Toor, Chen, Anklin, Wang, Feng, Gholami, Ling, Liu, Walter, Moghaddam, Kishore, Adamek, Mercado, Mallinson, Wandekar, Cagle, Ofek, Garrido, Lombriser, Mukha, Sun, Mohammad, Matak, Qian, Peswani, Janus, Yuan, Schelin, David, Garg, He, Duzhyi, Älgmyr, Lottaz, Li, Yadav, Xu, Chinien, Shivanna, Chuklin, Li, Spadine, Wolfe, Mohamed, Das, Dai, He, von Dincklage, Upadhyay, Maurya, Chi, Krause, Salama, Rabinovitch, M, Selvan, Dektiarev, Ghiasi, Guven, Gupta, Liu, Sharma, Shtacher, Paul, Akerlund, Aubet, Huang, Zhu, Zhu, Teixeira, Fritze, Bertolini, Marinescu, Bölle, Paulus, Gupta, Latkar, Chang, Sanders, Wilson, Wu, Tan, Thiet, Doshi, Lall, Mishra, Chen, Luong, Benjamin, Lee, Andrejczuk, Rabiej, Ranjan, Styrc, Yin, Simon,
  Harriott, Bansal, Robsky, Bacon, Greene, Mirylenka, Zhou, Sarvana, Goyal, Andermatt, Siegler, Horn, Israel, Pongetti, Chen, Selvatici, Silva, Wang, Tolins, Guu, Yogev, Cai, Agostini, Shah, Nguyen, Donnaile, Pereira, Friso, Stambler, Kurzrok, Kuang, Romanikhin, Geller, Yan, Jang, Lee, Fica, Malmi, Tan, Banica, Balle, Pham, Huang, Avram, Shi, Singh, Hidey, Ahuja, Saxena, Dooley, Potharaju, O'Neill, Gokulchandran, Foley, Zhao, Dusenberry, Liu, Mehta, Kotikalapudi, Safranek-Shrader, Goodman, Kessinger, Globen, Kolhar, Gorgolewski, Ibrahim, Song, Eichenbaum, Brovelli, Potluri, Lahoti, Baetu, Ghorbani, Chen, Crawford, Pal, Sridhar, Gurita, Mujika, Petrovski, Cedoz, Li, Chen, Santo, Goyal, Punjabi, Kappaganthu, Kwak, LV, Velury, Choudhury, Hall, Shah, Figueira, Thomas, Lu, Zhou, Kumar, Jurdi, Chikkerur, Ma, Yu, Kwak, Ähdel, Rajayogam, Choma, Liu, Barua, Ji, Park, Hellendoorn, Bailey, Bilal, Zhou, Khatir, Sutton, Rzadkowski, Macintosh, Shagin, Medina, Liang, Zhou, Shah, Bi, Dankovics, Banga, Lehmann, Bredesen,
  Lin, Hoffmann, Lai, Chung, Yang, Balani, Bražinskas, Sozanschi, Hayes, Alcalde, Makarov, Chen, Stella, Snijders, Mandl, Kärrman, Nowak, Wu, Dyck, Vaidyanathan, R, Mallet, Rudominer, Johnston, Mittal, Udathu, Christensen, Verma, Irving, Santucci, Elsayed, Davoodi, Georgiev, Tenney, Hua, Cideron, Leurent, Alnahlawi, Georgescu, Wei, Zheng, Scandinaro, Jiang, Snoek, Sundararajan, Wang, Ontiveros, Karo, Cole, Rajashekhar, Tumeh, Ben-David, Jain, Uesato, Datta, Bunyan, Wu, Zhang, Stanczyk, Zhang, Steiner, Naskar, Azzam, Johnson, Paszke, Chiu, Elias, Mohiuddin, Muhammad, Miao, Lee, Vieillard, Park, Zhang, Stanway, Garmon, Karmarkar, Dong, Lee, Kumar, Zhou, Evens, Isaac, Irving, Loper, Fink, Arkatkar, Chen, Shafran, Petrychenko, Chen, Jia, Levskaya, Zhu, Grabowski, Mao, Magni, Yao, Snaider, Casagrande, Palmer, Suganthan, Castaño, Giannoumis, Kim, Rybiński, Sreevatsa, Prendki, Soergel, Goedeckemeyer, Gierke, Jafari, Gaba, Wiesner, Wright, Wei, Vashisht, Kulizhskaya, Hoover, Le, Li, Iwuanyanwu, Liu, Ramirez,
  Khorlin, Cui, LIN, Wu, Aguilar, Pallo, Chakladar, Perng, Abellan, Zhang, Dasgupta, Kushman, Penchev, Repina, Wu, van~der Weide, Ponnapalli, Kaplan, Simsa, Li, Dousse, Yang, Piper, Ie, Pasumarthi, Lintz, Vijayakumar, Andor, Valenzuela, Lui, Paduraru, Peng, Lee, Zhang, Greene, Nguyen, Kurylowicz, Hardin, Dixon, Janzer, Choo, Feng, Zhang, Singhal, Du, McKinnon, Antropova, Bolukbasi, Keller, Reid, Finchelstein, Raad, Crocker, Hawkins, Dadashi, Gaffney, Franko, Bulanova, Leblond, Chung, Askham, Cobo, Xu, Fischer, Xu, Sorokin, Alberti, Lin, Evans, Dimitriev, Forbes, Banarse, Tung, Omernick, Bishop, Sterneck, Jain, Xia, Amid, Piccinno, Wang, Banzal, Mankowitz, Polozov, Krakovna, Brown, Bateni, Duan, Firoiu, Thotakuri, Natan, Geist, tan Girgin, Li, Ye, Roval, Tojo, Kwong, Lee-Thorp, Yew, Sinopalnikov, Ramos, Mellor, Sharma, Wu, Miller, Sonnerat, Vnukov, Greig, Beattie, Caveness, Bai, Eisenschlos, Korchemniy, Tsai, Jasarevic, Kong, Dao, Zheng, Liu, Yang, Zhu, Teh, Sanmiya, Gladchenko, Trdin, Toyama, Rosen, Tavakkol,
  Xue, Elkind, Woodman, Carpenter, Papamakarios, Kemp, Kafle, Grunina, Sinha, Talbert, Wu, Owusu-Afriyie, Du, Thornton, Pont-Tuset, Narayana, Li, Fatehi, Wieting, Ajmeri, Uria, Ko, Knight, Héliou, Niu, Gu, Pang, Li, Levine, Stolovich, Santamaria-Fernandez, Goenka, Yustalim, Strudel, Elqursh, Deck, Lee, Li, Levin, Hoffmann, Holtmann-Rice, Bachem, Arora, Koh, Yeganeh, Põder, Tariq, Sun, Ionita, Seyedhosseini, Tafti, Liu, Gulati, Liu, Ye, Chrzaszcz, Wang, Sethi, Li, Brown, Singh, Fan, Parisi, Stanton, Koverkathu, Choquette-Choo, Li, Lu, Ittycheriah, Shroff, Varadarajan, Bahargam, Willoughby, Gaddy, Desjardins, Cornero, Robenek, Mittal, Albrecht, Shenoy, Moiseev, Jacobsson, Ghaffarkhah, Rivière, Walton, Crepy, Parrish, Zhou, Farabet, Radebaugh, Srinivasan, van~der Salm, Fidjeland, Scellato, Latorre-Chimoto, Klimczak-Plucińska, Bridson, de~Cesare, Hudson, Mendolicchio, Walker, Morris, Mauger, Guseynov, Reid, Odoom, Loher, Cotruta, Yenugula, Grewe, Petrushkina, Duerig, Sanchez, Yadlowsky, Shen, Globerson, Webb,
  Dua, Li, Bhupatiraju, Hurt, Qureshi, Agarwal, Shani, Eyal, Khare, Belle, Wang, Tekur, Kale, Wei, Sang, Saeta, Liechty, Sun, Zhao, Lee, Nayak, Fritz, Vuyyuru, Aslanides, Vyas, Wicke, Ma, Eltyshev, Martin, Cate, Manyika, Amiri, Kim, Xiong, Kang, Luisier, Tripuraneni, Madras, Guo, Waters, Wang, Ainslie, Baldridge, Zhang, Pruthi, Bauer, Yang, Mansour, Gelman, Xu, Polovets, Liu, Cai, Chen, Sheng, Xue, Ozair, Angermueller, Li, Sinha, Wang, Wiesinger, Koukoumidis, Tian, Iyer, Gurumurthy, Goldenson, Shah, Blake, Yu, Urbanowicz, Palomaki, Fernando, Durden, Mehta, Momchev, Rahimtoroghi, Georgaki, Raul, Ruder, Redshaw, Lee, Zhou, Jalan, Li, Hechtman, Schuh, Nasr, Milan, Mikulik, Franco, Green, Nguyen, Kelley, Mahendru, Hu, Howland, Vargas, Hui, Bansal, Rao, Ghiya, Wang, Ye, Sarr, Preston, Elish, Li, Kaku, Gupta, Pasupat, Juan, Someswar, M., Chen, Amini, Fabrikant, Chu, Dong, Muthal, Buthpitiya, Jauhari, Hua, Khandelwal, Hitron, Ren, Rinaldi, Drath, Dabush, Jiang, Godhia, Sachs, Chen, Fan, Taitelbaum, Noga, Dai, Wang,
  Liang, Hamer, Ferng, Elkind, Atias, Lee, Listík, Carlen, van~de Kerkhof, Pikus, Zaher, Müller, Zykova, Stefanec, Gatsko, Hirnschall, Sethi, Xu, Ahuja, Tsai, Stefanoiu, Feng, Dhandhania, Katyal, Gupta, Parulekar, Pitta, Zhao, Bhatia, Bhavnani, Alhadlaq, Li, Danenberg, Tu, Pine, Filippova, Ghosh, Limonchik, Urala, Lanka, Clive, Sun, Li, Wu, Hongtongsak, Li, Thakkar, Omarov, Majmundar, Alverson, Kucharski, Patel, Jain, Zabelin, Pelagatti, Kohli, Kumar, Kim, Sankar, Shah, Ramachandruni, Zeng, Bariach, Weidinger, Vu, Andreev, He, Hui, Kashem, Subramanya, Hsiao, Hassabis, Kavukcuoglu, Sadovsky, Le, Strohman, Wu, Petrov, Dean, and Vinyals}]{geminiteam2024geminifamilyhighlycapable}
Gemini Team, Rohan Anil, Sebastian Borgeaud, Jean-Baptiste Alayrac, Jiahui Yu, Radu Soricut, Johan Schalkwyk, Andrew~M. Dai, Anja Hauth, Katie Millican, David Silver, Melvin Johnson, Ioannis Antonoglou, Julian Schrittwieser, Amelia Glaese, Jilin Chen, Emily Pitler, Timothy Lillicrap, Angeliki Lazaridou, and 1331 others. 2024.
\newblock \href {https://arxiv.org/abs/2312.11805} {Gemini: A family of highly capable multimodal models}.
\newblock \emph{Preprint}, arXiv:2312.11805.

\bibitem[{Team et~al.(2022)Team, Costa-jussà, Cross, Çelebi, Elbayad, Heafield, Heffernan, Kalbassi, Lam, Licht, Maillard, Sun, Wang, Wenzek, Youngblood, Akula, Barrault, Gonzalez, Hansanti, Hoffman, Jarrett, Sadagopan, Rowe, Spruit, Tran, Andrews, Ayan, Bhosale, Edunov, Fan, Gao, Goswami, Guzmán, Koehn, Mourachko, Ropers, Saleem, Schwenk, and Wang}]{nllbteam2022languageleftbehindscaling}
NLLB Team, Marta~R. Costa-jussà, James Cross, Onur Çelebi, Maha Elbayad, Kenneth Heafield, Kevin Heffernan, Elahe Kalbassi, Janice Lam, Daniel Licht, Jean Maillard, Anna Sun, Skyler Wang, Guillaume Wenzek, Al~Youngblood, Bapi Akula, Loic Barrault, Gabriel~Mejia Gonzalez, Prangthip Hansanti, and 20 others. 2022.
\newblock \href {https://arxiv.org/abs/2207.04672} {No language left behind: Scaling human-centered machine translation}.
\newblock \emph{Preprint}, arXiv:2207.04672.

\bibitem[{Touvron et~al.(2023)Touvron, Martin, Stone, Albert, Almahairi, Babaei, Bashlykov, Batra, Bhargava, Bhosale, Bikel, Blecher, Ferrer, Chen, Cucurull, Esiobu, Fernandes, Fu, Fu, Fuller, Gao, Goswami, Goyal, Hartshorn, Hosseini, Hou, Inan, Kardas, Kerkez, Khabsa, Kloumann, Korenev, Koura, Lachaux, Lavril, Lee, Liskovich, Lu, Mao, Martinet, Mihaylov, Mishra, Molybog, Nie, Poulton, Reizenstein, Rungta, Saladi, Schelten, Silva, Smith, Subramanian, Tan, Tang, Taylor, Williams, Kuan, Xu, Yan, Zarov, Zhang, Fan, Kambadur, Narang, Rodriguez, Stojnic, Edunov, and Scialom}]{touvron2023llama2openfoundation}
Hugo Touvron, Louis Martin, Kevin Stone, Peter Albert, Amjad Almahairi, Yasmine Babaei, Nikolay Bashlykov, Soumya Batra, Prajjwal Bhargava, Shruti Bhosale, Dan Bikel, Lukas Blecher, Cristian~Canton Ferrer, Moya Chen, Guillem Cucurull, David Esiobu, Jude Fernandes, Jeremy Fu, Wenyin Fu, and 49 others. 2023.
\newblock \href {https://arxiv.org/abs/2307.09288} {Llama 2: Open foundation and fine-tuned chat models}.
\newblock \emph{Preprint}, arXiv:2307.09288.

\bibitem[{Wei et~al.(2023)Wei, Wang, Schuurmans, Bosma, Ichter, Xia, Chi, Le, and Zhou}]{wei2023chainofthoughtpromptingelicitsreasoning}
Jason Wei, Xuezhi Wang, Dale Schuurmans, Maarten Bosma, Brian Ichter, Fei Xia, Ed~Chi, Quoc Le, and Denny Zhou. 2023.
\newblock \href {https://arxiv.org/abs/2201.11903} {Chain-of-thought prompting elicits reasoning in large language models}.
\newblock \emph{Preprint}, arXiv:2201.11903.

\bibitem[{Xu et~al.(2024)Xu, Fei, Pan, Liu, Lee, and Hsu}]{xu2024faithfullogicalreasoningsymbolic}
Jundong Xu, Hao Fei, Liangming Pan, Qian Liu, Mong-Li Lee, and Wynne Hsu. 2024.
\newblock \href {https://arxiv.org/abs/2405.18357} {Faithful logical reasoning via symbolic chain-of-thought}.
\newblock \emph{Preprint}, arXiv:2405.18357.

\bibitem[{Ying et~al.(2024)Ying, Wu, Geng, Wang, Lin, and Chen}]{ying2024leanworkbooklargescalelean}
Huaiyuan Ying, Zijian Wu, Yihan Geng, Jiayu Wang, Dahua Lin, and Kai Chen. 2024.
\newblock \href {https://arxiv.org/abs/2406.03847} {Lean workbook: A large-scale lean problem set formalized from natural language math problems}.
\newblock \emph{Preprint}, arXiv:2406.03847.

\bibitem[{Zhang et~al.(2024)Zhang, Zhang, Li, and Xing}]{zhang2024evaluatingstepbystepreasoningsymbolic}
Yi-Fan Zhang, Hanlin Zhang, Li~Erran Li, and Eric Xing. 2024.
\newblock \href {https://arxiv.org/abs/2212.08686} {Evaluating step-by-step reasoning through symbolic verification}.
\newblock \emph{Preprint}, arXiv:2212.08686.

\bibitem[{Zheng et~al.(2022)Zheng, Han, and Polu}]{zheng2022minif2fcrosssystembenchmarkformal}
Kunhao Zheng, Jesse~Michael Han, and Stanislas Polu. 2022.
\newblock \href {https://arxiv.org/abs/2109.00110} {Minif2f: a cross-system benchmark for formal olympiad-level mathematics}.
\newblock \emph{Preprint}, arXiv:2109.00110.

\end{thebibliography}

\clearpage

\appendix

\section{Appendix}

\begin{table}[!ht]
    \centering
    \begin{tabular}{|l|l|l|l|l|l|}
    \hline
    \multicolumn{6}{|c|}{Translation Key 1 (25214)} \\ \hline
    \multicolumn{1}{|c|}{~} & \multicolumn{1}{c|}{Testing Dataset} & \multicolumn{1}{c|}{Accuracy (\%)} & \multicolumn{1}{c|}{Total Queries} & \multicolumn{1}{c|}{Correct} & \multicolumn{1}{c|}{Incorrect} \\ \hline
    Seen   & Lean (500) - Benchmark& 76.66666667 & 500 & 383.3333333 & 116.6666667 \\ \hline
    ~      & 1                      & 76.2       & 500 & 381        & 119         \\ \hline
    ~      & 2                      & 74.4       & 500 & 372        & 128         \\ \hline
    ~      & 3                      & 79.4       & 500 & 397        & 103         \\ \hline
    Seen   & Translated (500)& 96.4       & 500 & 482        & 18          \\ \hline
    ~      & 1                      & 96.4       & 500 & 482        & 18          \\ \hline
    ~      & 2                      & 97         & 500 & 485        & 15          \\ \hline
    ~      & 3                      & 95.8       & 500 & 479        & 21          \\ \hline
    Unseen & Lean (200)& 100        & 200 & 200        & 0           \\ \hline
    ~      & 1                      & 100        & 200 & 200        & 0           \\ \hline
    ~      & 2                      & 100        & 200 & 200        & 0           \\ \hline
    ~      & 3                      & 100        & 200 & 200        & 0           \\ \hline
    Unseen & Translated (200)& 97.66666667& 200 & 195.3333333& 4.666666667 \\ \hline
    ~      & 1                      & 98         & 200 & 196        & 4           \\ \hline
    ~      & 2                      & 98         & 200 & 196        & 4           \\ \hline
    ~      & 3                      & 97         & 200 & 194        & 6           \\ \hline
    \multicolumn{6}{|c|}{Translation Key 1 (20000)} \\ \hline
    \multicolumn{1}{|c|}{~} & \multicolumn{1}{c|}{Testing Dataset} & \multicolumn{1}{c|}{Accuracy (\%)} & \multicolumn{1}{c|}{Total Queries} & \multicolumn{1}{c|}{Correct} & \multicolumn{1}{c|}{Incorrect} \\ \hline
    Seen   & Lean (500) - Benchmark& 76.66667   & 500 & 383.3333   & 116.6667    \\ \hline
    ~      & 1                      & 76.2       & 500 & 381        & 119         \\ \hline
    ~      & 2                      & 74.4       & 500 & 372        & 128         \\ \hline
    ~      & 3                      & 79.4       & 500 & 397        & 103         \\ \hline
    Seen   & Translated (500)& 96.4       & 500 & 482        & 18          \\ \hline
    ~      & 1                      & 96.4       & 500 & 482        & 18          \\ \hline
    ~      & 2                      & 97         & 500 & 485        & 15          \\ \hline
    ~      & 3                      & 95.8       & 500 & 479        & 21          \\ \hline
    Unseen & Lean (200)& 100        & 200 & 200        & 0           \\ \hline
    ~      & 1                      & 100        & 200 & 200        & 0           \\ \hline
    ~      & 2                      & 100        & 200 & 200        & 0           \\ \hline
    ~      & 3                      & 100        & 200 & 200        & 0           \\ \hline
    Unseen & Translated (200)& 97.66667   & 200 & 195.3333   & 4.666667    \\ \hline
    ~      & 1                      & 98         & 200 & 196        & 4           \\ \hline
    ~      & 2                      & 98         & 200 & 196        & 4           \\ \hline
    ~      & 3                      & 97         & 200 & 194        & 6           \\ \hline
    \end{tabular}
    \caption{Summary table for Translation Key 1 model evaluation results. The top of each testing dataset shows the overall average results across three runs. (Part 1/2)}
    \label{tab:performance-translationkey1}
\end{table}

\clearpage

\begin{table}[!ht]
    \centering
    \begin{tabular}{|l|l|l|l|l|l|}
    \hline
    \multicolumn{6}{|c|}{Translation Key 1 (10000)} \\ \hline
    \multicolumn{1}{|c|}{~} & \multicolumn{1}{c|}{Testing Dataset} & \multicolumn{1}{c|}{Accuracy (\%)} & \multicolumn{1}{c|}{Total Queries} & \multicolumn{1}{c|}{Correct} & \multicolumn{1}{c|}{Incorrect} \\ \hline
    Seen   & Lean (500) - Benchmark& 76.93333333& 500 & 384.6666667& 115.3333333\\ \hline
    ~      & 1                      & 80         & 500 & 400         & 100         \\ \hline
    ~      & 2                      & 74.4       & 500 & 372         & 128         \\ \hline
    ~      & 3                      & 76.4       & 500 & 382         & 118         \\ \hline
    Seen   & Translated (500)& 96.86666667& 500 & 484.3333333& 15.66666667\\ \hline
    ~      & 1                      & 97.2       & 500 & 486         & 14          \\ \hline
    ~      & 2                      & 97.2       & 500 & 486         & 14          \\ \hline
    ~      & 3                      & 96.2       & 500 & 481         & 19          \\ \hline
    Unseen & Lean (200)& 99.66666667& 200 & 199.3333333 & 0.666666667\\ \hline
    ~      & 1                      & 99.5       & 200 & 199         & 1           \\ \hline
    ~      & 2                      & 99.5       & 200 & 199         & 1           \\ \hline
    ~      & 3                      & 100        & 200 & 200         & 0           \\ \hline
    Unseen & Translated (200)& 97.33333333& 200 & 194.6666667& 5.333333333 \\ \hline
    ~      & 1                      & 97.5       & 200 & 195         & 5           \\ \hline
    ~      & 2                      & 97.5       & 200 & 195         & 5           \\ \hline
    ~      & 3                      & 97         & 200 & 194         & 6           \\ \hline
    \end{tabular}
    \caption{Summary table for Translation Key 1 model evaluation results.  The top of each testing dataset shows the overall average results across three runs. (Part 2/2)}
    \label{tab:performance-translationkey2}
\end{table}

\begin{table}[!ht]
    \centering
    \begin{tabular}{|l|l|l|l|l|l|}
    \hline
    \multicolumn{6}{|c|}{Translation Key 2 (25214)} \\ \hline
    \multicolumn{1}{|c|}{~} & \multicolumn{1}{c|}{Testing Dataset} & \multicolumn{1}{c|}{Accuracy (\%)} & \multicolumn{1}{c|}{Total Queries} & \multicolumn{1}{c|}{Correct} & \multicolumn{1}{c|}{Incorrect} \\ \hline
    Seen   & Lean (500) - Benchmark & 74.06666667& 500 & 370.3333333 & 129.6666667 \\ \hline
    ~      & 1                      & 74.6       & 500 & 373         & 127         \\ \hline
    ~      & 2                      & 72         & 500 & 360         & 140         \\ \hline
    ~      & 3                      & 75.6       & 500 & 378         & 122         \\ \hline
    Seen   & Translated (500)& 94.2       & 500 & 471         & 29          \\ \hline
    ~      & 1                      & 92.6       & 500 & 463         & 37          \\ \hline
    ~      & 2                      & 96.2       & 500 & 481         & 19          \\ \hline
    ~      & 3                      & 93.8       & 500 & 469         & 31          \\ \hline
    Unseen & Lean (200)& 64.16666667 & 200 & 128.3333333 & 71.66666667 \\ \hline
    ~      & 1                      & 64         & 200 & 128         & 72          \\ \hline
    ~      & 2                      & 63.5       & 200 & 127         & 73          \\ \hline
    ~      & 3                      & 65         & 200 & 130         & 70          \\ \hline
    Unseen & Translated (200)& 89         & 200 & 178         & 22          \\ \hline
    ~      & 1                      & 91         & 200 & 182         & 18          \\ \hline
    ~      & 2                      & 88         & 200 & 176         & 24          \\ \hline
    ~      & 3                      & 88         & 200 & 176         & 24          \\ \hline
    \end{tabular}
    \caption{Summary table for Translation Key 2 model evaluation results.  The top of each testing dataset shows the overall average results across three runs.}
    \label{tab:performance-translationkey2}
\end{table}

\clearpage

\begin{table}[!ht]
    \centering
    \begin{tabular}{|l|l|l|l|l|l|}
    \hline
    \multicolumn{6}{|c|}{Base GPT-4o - Translation Key 1} \\ \hline
    \multicolumn{1}{|c|}{~} & \multicolumn{1}{c|}{Testing Dataset} & \multicolumn{1}{c|}{Accuracy (\%)} & \multicolumn{1}{c|}{Total Queries} & \multicolumn{1}{c|}{Correct} & \multicolumn{1}{c|}{Incorrect} \\ \hline
    Seen   & Lean (500) - Benchmark& 73.53333 & 500 & 367.66667 & 132.33333 \\ \hline
    ~      & 1                               & 73.4     & 500 & 367       & 133       \\ \hline
    ~      & 2                               & 70.4     & 500 & 352       & 148       \\ \hline
    ~      & 3                               & 76.8     & 500 & 384       & 116       \\ \hline
    Seen& Translated (500)& 23.33333 & 500 & 117       & 383       \\ \hline
    ~      & 1                               & 25.2     & 500 & 126       & 374       \\ \hline
    ~      & 2                               & 21.6     & 500 & 108       & 392       \\ \hline
    ~      & 3                               & 23.2     & 500 & 116       & 384       \\ \hline
    Unseen & Lean (200)& 91.83333 & 200 & 183.66667 & 16.33333  \\ \hline
    ~      & 1                               & 92       & 200 & 184       & 16        \\ \hline
    ~      & 2                               & 91       & 200 & 182       & 18        \\ \hline
    ~      & 3                               & 92.5     & 200 & 185       & 15        \\ \hline
    Unseen & Translated (200)& 4        & 200 & 8         & 192       \\ \hline
    ~      & 1                               & 4        & 200 & 8         & 192       \\ \hline
    ~      & 2                               & 4        & 200 & 8         & 192       \\ \hline
    ~      & 3                               & 4        & 200 & 8         & 192       \\  
    \hline
    \multicolumn{6}{|c|}{Base GPT-4o - Translation Key 2} \\ \hline
    \multicolumn{1}{|c|}{~} & \multicolumn{1}{c|}{Testing Dataset} & \multicolumn{1}{c|}{Accuracy (\%)} & \multicolumn{1}{c|}{Total Queries} & \multicolumn{1}{c|}{Correct} & \multicolumn{1}{c|}{Incorrect} \\ \hline
    Seen   & Lean (500) - Benchmark& 73.53333   & 500 & 367.66667   & 132.33333    \\ \hline
    ~      & 1                             & 73.4       & 500 & 367         & 133          \\ \hline
    ~      & 2                             & 76.2       & 500 & 381         & 119          \\ \hline
    ~      & 3                             & 71         & 500 & 355         & 145          \\ \hline
    Seen& Translated (500)& 26.66667   & 500 & 133.33333   & 366.66667    \\ \hline
    ~      & 1                             & 27.4       & 500 & 137         & 363          \\ \hline
    ~      & 2                             & 28.2       & 500 & 141         & 359          \\ \hline
    ~      & 3                             & 24.4       & 500 & 122         & 378          \\ \hline
    Unseen & Lean (200)& 90         & 200 & 180         & 20           \\ \hline
    ~      & 1                             & 90         & 200 & 180         & 20           \\ \hline
    ~      & 2                             & 90         & 200 & 180         & 20           \\ \hline
    ~      & 3                             & 90         & 200 & 180         & 20           \\ \hline
    Unseen& Translated (200)& 4.16667    & 200 & 8.33333     & 191.66667    \\ \hline
    ~      & 1                             & 3.5        & 200 & 7           & 193          \\ \hline
    ~      & 2                             & 4.5        & 200 & 9           & 191          \\ \hline
    ~      & 3                             & 4.5        & 200 & 9           & 191          \\ \hline
    \end{tabular}
    \caption{Summary table for Base GPT-4o for Translation Key 1 and Translation Key 2.  The top of each testing dataset shows the overall average results across three runs.}
    \label{tab:best-gpt4o}
\end{table}

\clearpage

\begin{figure*}
    \centering
    \includegraphics[scale=0.95]{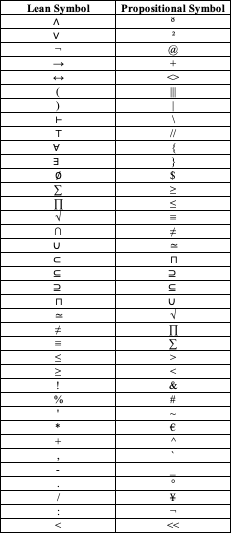}
    \caption{Mapping between Lean's logical symbols and their corresponding representations in our custom propositional language. (Part 1/2)}
\end{figure*}
\clearpage

\begin{figure*}
    \centering
    \includegraphics[scale=0.95]{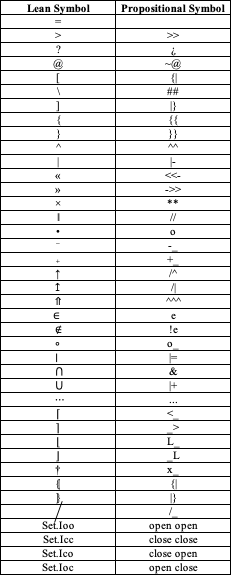}
    \caption{Mapping between Lean's logical symbols and their corresponding representations in our custom propositional language. (Part 2/2)}
\end{figure*}

\end{document}